\documentclass[runningheads]{llncs}
 
\usepackage{eccv}

\usepackage{eccvabbrv}

\usepackage{graphicx}
\usepackage{booktabs}

\usepackage[accsupp]{axessibility}

\usepackage[breaklinks,colorlinks,citecolor=eccvblue]{hyperref}

\begin{document}

\title{MoDiPO: text-to-motion alignment via AI-feedback-driven Direct Preference Optimization} 

\titlerunning{MoDiPO}

\author{Massimiliano Pappa*
\and
Luca Collorone*
\and
Giovanni Ficarra
\and
Indro Spinelli
\and
Fabio Galasso
}

\authorrunning{M.~Pappa \etal}

\institute{
Sapienza University of Rome
}

\maketitle
\def\thefootnote{*}\footnotetext{Authors contributed equally.}

\begin{abstract}
Diffusion Models have revolutionized the field of human motion generation by offering exceptional generation quality and fine-grained controllability through natural language conditioning.
Their inherent stochasticity, that is the ability to generate various outputs from a single input, is key to their success.
However, this diversity should not be unrestricted, as it may lead to unlikely generations. Instead, it should be confined within the boundaries of text-aligned and realistic generations.
To address this issue, we propose MoDiPO (Motion Diffusion DPO), a novel methodology that leverages Direct Preference Optimization (DPO) to align text-to-motion models. We streamline the laborious and expensive process of gathering human preferences needed in DPO by leveraging AI feedback instead.
This enables us to experiment with novel DPO strategies, using both online and offline generated motion-preference pairs. To foster future research we contribute with a motion-preference dataset which we dub Pick-a-Move.
We demonstrate, both qualitatively and quantitatively, that our proposed method yields significantly more realistic motions. In particular, MoDiPO substantially improves Frechet Inception Distance (FID) while retaining the same RPrecision and Multi-Modality performances.
\end{abstract}    
\section{Introduction}
Synthesizing realistic human behavior is a fundamental challenge in computer animation, holding significant relevance in gaming, film, and virtual reality contexts. Recent advancements have shifted towards text-driven generation methods, aiming to bypass time-consuming traditional modeling processes requiring specialized equipment and artists' domain knowledge. Instead, leveraging natural language descriptions provides sufficient semantic details to describe generic human motion. One key advantage of text-driven generative models is their probabilistic nature, allowing for multiple diverse generations from the same prompt, each characterized by unique nuances. Yet, a high level of generative diversity also introduces the risk of producing motions that are either unrealistic or deviate significantly from the intended ones. To our knowledge, no research has addressed this aspect so far.

\begin{figure}
\includegraphics[width=\textwidth]{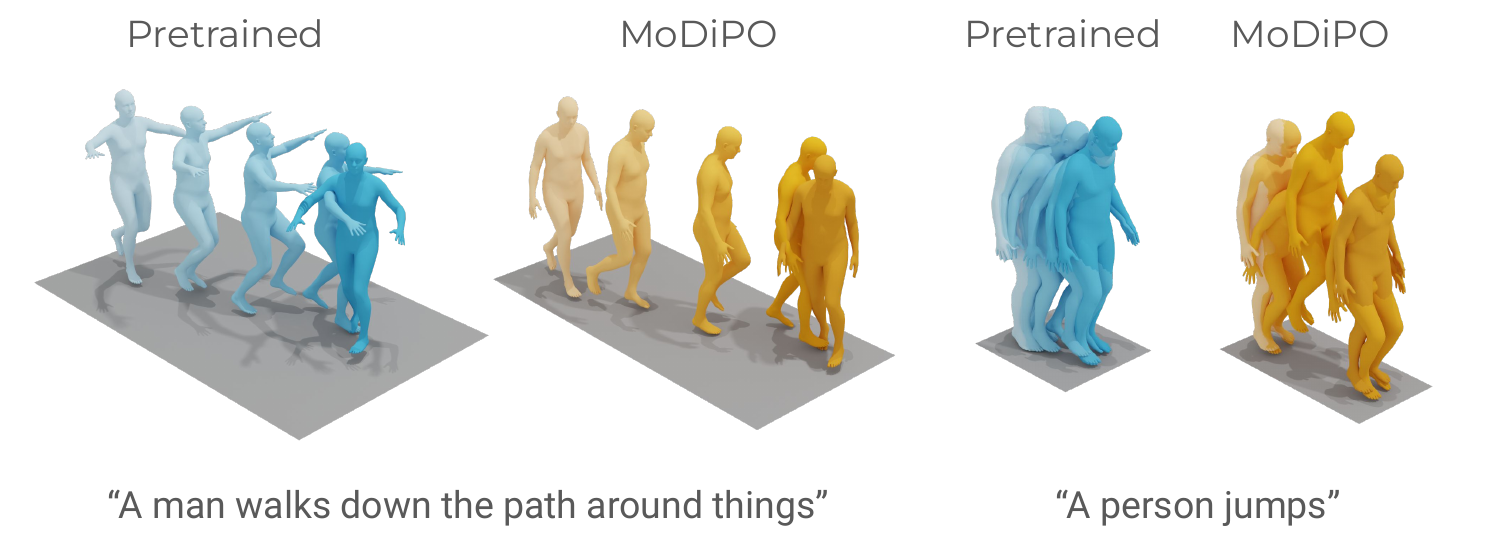}
\caption{By fine-tuning the models on a synthetic preferential dataset, we enhance the realism of generated motions. The left showcases MoDiPO's motion walking naturally, avoiding specific areas. Similarly, on the right, the jump appears more natural compared to the unaligned model's generation.}
\end{figure}

Recently, \cite{christiano2017deep,bai2022constitutional} introduced alignment methodologies allowing users to instill desired behaviors in pre-trained generative models by leveraging curated sets of human preferences. This approach has succeeded in various tasks, including question answering \cite{menick2022teaching}, dialogue support, and even image generation \cite{wu2023human}. Of particular interest, synthetic or AI-generated feedback not only circumvents the time and cost associated with traditional human-annotated data collection methods but also scales up the dimensions of preference datasets.
Despite substantial efforts in developing text-driven 3D human motion generative models, our research stands out as the first to explore alignment techniques for this task.

In order to exploit the creativity of the generative pipeline, we would like to bias our model toward high-quality motions, avoiding altogether regions of the learned latent manifold that lead to unfeasible motions. We base our work on the adaptation of Direct Preference Optimization \cite{rafailov2023direct} for DDPMs \cite{wallace2023diffusion} to directly optimize the models on the preferential data.
For this purpose, we introduce ``Pick-a-Move'', a dataset designed to capture human preferences for motion. To achieve this, we utilize motions generated by SOTA text-to-motion models. These motions are then paired with preference scores provided by a text-to-motion retrieval model \cite{Guo22humanml3d, petrovich2023tmr}. These models operate within a shared latent space, where related text and motion pairs are positioned closer together. This shared space allows the retrieval model to express preferences for specific text-motion combinations.
We use Pick-a-Move to align text-driven diffusion-based generative models \cite{tevet2023mdm,chen2023mld} for motion synthesis.
Specifically, MLD~\cite{chen2023mld} is selected as it strikes a compromise between efficiency and realism, the latter witnessed by the Frechet Inception Distance (FID).

This approach might penalize realistic and high-quality motions that do not perfectly align with the input text. Note that training models solely on pre-defined motion-preference pairs can restrict the model's diversity. For this reason, Pick-a-Move provides more flexibility in training by offering up to $N$ motion options for each descriptive text. These motions are ranked from most to least preferred. This allows training on pairs of text and motion where the selection of \textit{winner} and \textit{loser} motion can be adjusted during the training process.

Additionally, we exploit the availability of ground truth motions associated with the input text by selecting these motions as winners with a low probability: this mitigates both the risk of overfitting on training data and the artifacts or motion-text misalignments introduced by training on generated sequences.
As a result, this supervised fine-tuning element within our pipeline further enhances the quality of the generated outputs.

Pioneering the combination of text-driven motion generation and alignment, we perform extensive evaluations on two datasets with two different models. We achieve improvements in FID of up to $39\%$ while retaining consistent RPrecision and MultiModality performances.
To summarize, our contributions are the following.
\begin{itemize}
    \item We propose the first DPO-based framework for aligning text-to-motion generative models.
    \item We introduce Pick-a-Move, a motion-preference dataset annotated leveraging synthetic feedback.
    \item We conduct an extensive experimental evaluation and ablation study to devise the best practices for a successful alignment in text-to-motion generation.
\end{itemize}

\section{Related Works}

\subsection{Aligning Models with Human Feedback}
\label{sub2:align}
Early applications of models alignment employed a complex reinforcement learning pipeline, requiring preferential data built through human feedback (RLHF) \cite{christiano2017deep}. This technique has been applied to text generation tasks such as summarization \cite{stiennon2020learning}. Advancements in the reinforcement learning pipeline by InstructGPT \cite{ouyang2022training}, such as the use of Proximal Policy Optimization \cite{schulman2017proximal}, have enabled the finetuning of models for tasks requiring evidence-supported question answering \cite{menick2022teaching} and dialogue support \cite{glaese2022improving}.
Several solutions have been proposed to simplify this complex pipeline: Implicit Language Q-Learning (ILQL) \cite{snell2022offline} utilizes Q-learning to learn value functions that guide language model generation to maximize user-specified utility functions. Similarly, HIVE \cite{zhang2023hive} leverages offline reinforcement learning for instruction editing alignment.
Direct Preference Optimization (DPO) \cite{rafailov2023direct} offers a reinforcement learning-free approach. It optimizes the model directly using a supervised classification objective on the preferential data, eliminating the need for an explicit reward model. This DPO approach has inspired several RL-free methods for text-to-image alignment\cite{lee2023aligning}  including Diffusion-DPO \cite{wu2023human}. Diffusion-DPO specializes the DPO objective for DDPMs \cite{ho2020denoising}, a popular choice for text-driven human motion generative models. For this reason and its simplicity, we have adopted this latest approach.

\subsection{Aligning Models with Synthetic Feedback}
Recent works have focused on the possibility of replacing human annotators with Synthetic Feedback (aka AI Feedback) to avoid the high costs and effort required to collect high-quality datasets.
One approach involves self-supervision (self-improvement), as demonstrated in \cite{bai2022constitutional}. In this method, human ``supervision" only defines a list of prompts, that the generative model should respect. The model then self-evaluates the generated sentences by computing a reward based on these human-specified guidelines. This synthetically labeled dataset is then used to train a preference model, which a policy aims to maximize using an RL algorithm.
The approach in \cite{kim2023aligning} utilizes synthetic ranking by using different models with varying levels of complexity to generate sentences based on the same input prompt, where the outputs from models with higher parameter counts, which tend to generate better content, are assigned higher rewards and, thus, a higher rank.
Similar techniques have been applied to text-to-image generation, as seen in \cite{black2023training}. Additionally, \cite{lee2023rlaif} demonstrated through extensive experiments that RLAIF (Reinforcement Learning from AI Feedback) can be comparable to, or even outperform, RLHF.
For these reasons, we utilize synthetic feedback in our pipeline. We leverage a retrieval model that measures the alignment between the produced motion and the input text to rank the generated motions.

\subsection{Text-to-Motion}
Early approaches like TEMOS \cite{petrovich22temos} directly used VAEs to generate motion based on textual conditioning. Currently, two leading generation strategies dominate. Autoregressive methods \cite{Radford2018gpt}, such as MotionGPT \cite{jiang2024motiongpt} and T2M-GPT \cite{zhang2023t2mgpt}, and Denoising Diffusion Probababilistic Models (DDPMs) \cite{ho2020denoising}. The latter category includes models like MDM \cite{tevet2023mdm} and MLD \cite{chen2023mld}. MDM applies DDPM directly on raw motion data to capture the motion-text relationship for a specific number of frames. MLD, instead, improves performances by utilizing a two-step approach: First, they learn a compact motion latent space using VAEs; then, they apply a diffusion process on the learned latent space leveraging Latent DDPMs \cite{rombach2022high}.
Our focus lies on improving DDPM-based models, specifically MLD, due to their significantly reduced training and inference times compared to other Diffusion-based approaches.
Furthermore, Retrieval-Augmented Generation \cite{lewis2020retrieval, zhang2023remo} demonstrates performance improvements when conditioning motion generation on a sample retrieved from the entire training set. However, this approach requires storing and searching a large database, the quality of which significantly impacts the generation. Conversely, aligning a model does not introduce additional computational or spatial complexity.

\section{Background}
Given our aim to enhance pre-trained motion diffusion models using the DPO strategy, in this section we offer an overview of motion DDPMs, and their latent counterparts. Notice that in our work we follow \cite{Guo22humanml3d} and represent a motion sequence as $\mathbf{p}= \{\mathbf{p}_k\}_{k=1}^{F}\in \mathbb{R}^{F\times V}$ where $F$ is the number of frames and $V$ represents the pose vector's features.

In Sec.~\ref{sub3:vae} we present an overview on latent human representation, which is a mandatory requirement for latent diffusion models. In Sec.~\ref{sub3:ddpm} and in Sec.~\ref{sub3:latentddpm} we briefly describe the DDPMs and latent-DDPMs frameworks.

\subsection{Latent Human Representation}
\label{sub3:vae}

VAEs consist of an encoder-decoder generative architecture trained to minimize the reconstruction error. 
We employ it to reduce the pose vector $V$ dimensionality, projecting onto a manifold of feasible poses.
In this framework, the encoder network, parameterized by $\phi$, generates lower-dimensional embeddings $\mathbf{z} \in \mathbb{R}^{n \times D}$ as outputs based on the input poses $\mathbf{p} = \{{\mathbf{p}_k}\}_{k=1}^{F}$.
Following VAE literature \cite{kingma2013auto}, $q_\phi(\mathbf{z} \vert \mathbf{p})$ approximates the true posterior distribution of the latent space with a multivariate Gaussian with a diagonal covariance structure:
\begin{equation}
q_\phi(\mathbf{z} \vert \mathbf{p}) = \mathcal{N}(\mathbf{z} \vert \mu_\phi(\mathbf{p}), \sigma^2_\phi(\mathbf{p})\mathbf{I}),
\end{equation}

where $\mu_\phi(\cdot)$ is the mean and $\sigma_\phi(\cdot)$ the standard deviation obtained as output from the encoder.
We sample from the approximated posterior $\mathbf{z}_i \sim q_\phi(\mathbf{z}\vert \mathbf{p}_i)$ using:
\begin{equation}
\label{eq:sampling}
\mathbf{z}_i = \boldsymbol{\mu}_i + \boldsymbol{\sigma}_{i}^2 \odot \boldsymbol{\rho},
\end{equation}
where $\mathbf{z}_i$ is a one-dimensional vector of size $D$, and $\boldsymbol{\rho}$ is sampled from a standard multivariate Gaussian distribution. The transformer decoder~\cite{chen2023mld, petrovich22temos} parametrized by $\theta$ maps the sampled values back to body pose representation $p_\theta(\mathbf{p} | \mathbf{z})$ which may be futher processed into body meshes with the differentiable SMPL model for visualization purposes. The network parameters are obtained by optimizing the Evidence Lower Bound (ELBO) objective, as described in~\cite{kingma2013auto}.

\subsection{DDPMs}
\label{sub3:ddpm}
Denoising Diffusion Probabilistic Models (DDPMs) \cite{ho2020denoising} work in two stages. First, they gradually add noise to the input data through a \textbf{forward process}, which is then corrupted according to a variance schedule $\beta_t \in (0,1)$ for $t \in \{1,\ldots,T\}$, transforming any data distribution $q(x_0)$ into a simple prior (e.g. Gaussian).
This forward process can be represented as
\begin{equation}
\label{eq:reverse1}
    q(x_t | x_{t-1})=\mathcal{N}(x_t;\sqrt{1-\beta_t}x_{t-1},\beta_t\mathbb{I})
\end{equation}

Then, a separate \textbf{reverse process} also dubbed as \textit{denoising} process, removes this noise, effectively cleaning the data, and is formulated as
\begin{equation}
\label{eq:reverse2}
    p_{\theta}(x_{t-1}|x_t) = \mathcal{N}\big(x_{t-1}; \mu_{\theta}(x_t,t), \beta_t\mathbb{I}\big)
\end{equation}
where $\mu_{\theta}(x_t,t)$ being the trained deep neural network that estimates the forward process posterior mean, a network that effectively acts as a denoiser.\newline
The loss term to optimize is thus
\begin{equation}
Loss := \mathbb{E}_{\epsilon \sim \mathcal{N}(0,\mathbf{I}),t} [\| \epsilon - \epsilon(\mathbf{x}_t, t) \|_2^2] \,.
\label{eq:loss_ddpm}
\end{equation}

\subsection{Latent DDPMs}
\label{sub3:latentddpm}
We base our analysis on the latent-DDPM frameworks ~\cite{rombach2022high,chen2023mld} where the diffusion process occurs on a condensed, low-dimensional motion latent space.
Contrary to the simple DDPMs that works on raw features, here the approximated posterior $g(\mathbf{z}_t \vert \mathbf{z}_{t-1})$ gradually converts latent representations $\mathbf{z}_{0}=\mathbf{z}$ into random noise $\mathbf{z}_{T}$ in $T$ timesteps.
Then, the reverse process gradually refines the noised vector to a suitable latent representation $\mathbf{z}_{0}$.\newline
Following \cite{chen2023mld, rombach2022high, ho2020denoising, dhariwal2021diffusion}, we use the notation $\{\mathbf{z}_t\}^T_{t=0}$ to denote the sequence of noised latent vectors, with $\mathbf{z}_{t-1} = \epsilon_\psi(\mathbf{z}_t, t)$ representing the denoising operation at time step $t$.
Here, $\epsilon_\psi$ refers to a denoising autoencoder trained to predict the denoised version of its input.
The forward and reverse processes are identical to those described in equations \ref{eq:reverse1} and \ref{eq:reverse2}, but they are directly applied to the latent vectors of the VAE. \newline
At inference, $\epsilon_\psi(\mathbf{z}_t, t)$ predicts $\hat{\mathbf{z}}_0$ through a series of $T$ iterative denoising steps and the frozen decoder translates $\hat{\mathbf{z}}_0$ into poses $\mathbf{P}$ in a single forward pass.

\section{Methodology}
\begin{figure}
    \centering
    \includegraphics[width=\textwidth]{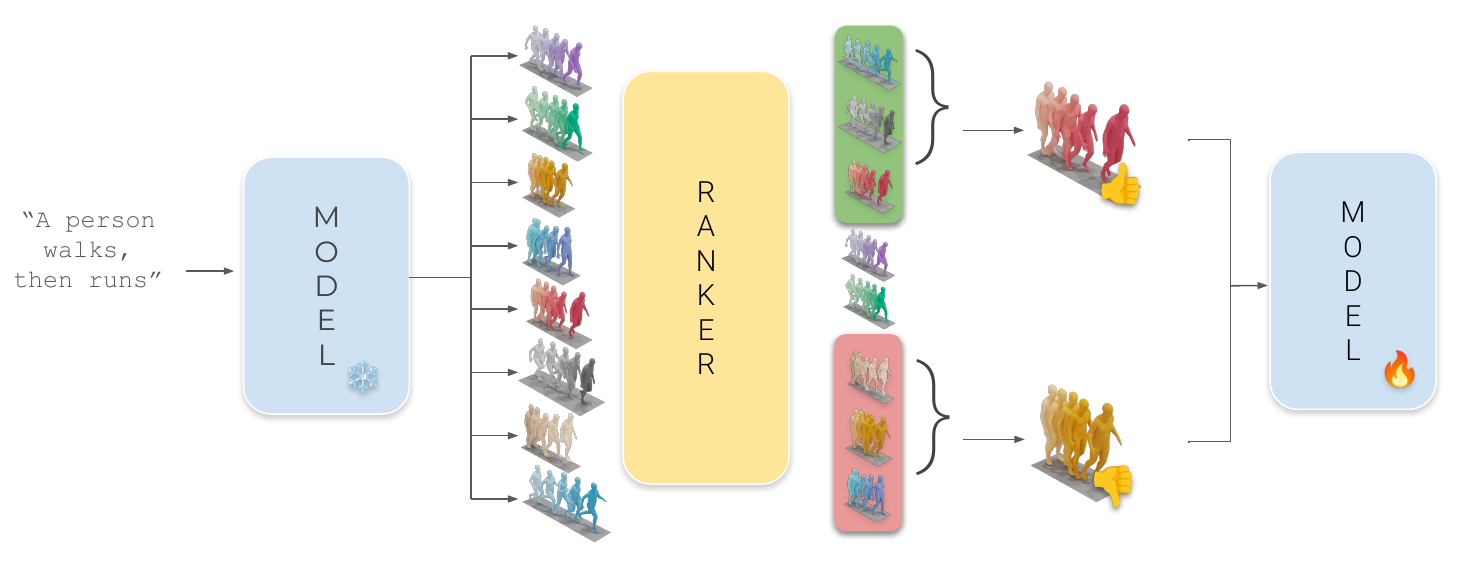}
    \caption{MoDiPO Schematics: Starting with the input prompt, we generate a winner-loser pair, which constitutes a sample in our preferential dataset. To do so, the reference model produces $K$ generations based on the same input prompt. These generations are then ranked by the ranker model according to their relevance with the textual input. From these rankings, we select both a set of winners and a set of losers. Finally, we sample from these sets to determine the final pair. This pair is then used to refine the unfrozen target model using DPO.}
    \label{fig:pictorial}
\end{figure}

In this section, we describe our approach to DPO-based text-to-motion alignment.
In Sec. \ref{sub4:dataset} we introduce a synthetic dataset made of motions, generated starting from an input prompt, motions that are also paired with AI-preferences. This dataset is then used by the alignment framework described in Sec. \ref{sub4:mot_alig}.

\subsection{Motion Preferences Dataset}
\label{sub4:dataset}
Unlike text and images, motion data often suffers from scarcity. At the same time, a dataset with motion preferences does not exist. To address this lack, we propose a synthetic dataset named Pick-a-Move, PaM for short.
To make PaM we generate $K$ motion, with both MLD and MDM, corresponding to a conditioning text $c$. Then, leveraging this set of $N=2K$ motion sequences $m=\{m_k\}^N_{k=1}$, we employ TMR \cite{petrovich2023tmr} and the model proposed by Guo \cite{Guo22humanml3d} to compute scores $s=\{s_k\}^N_{k=1}$ between the motion description and the generated motions. We use $s$ to order $m$ into $m^*=\{m^*_k\}^N_{k=1}$, representing a list where motions are presented from the most coherent with $c$, namely \textit{winner}, to least, \textit{loser}. This list of ranked synthetic motion sequences mimics human preferences in ordering motions from least to most accurate according to the prompt they were generated from. Thanks to this dataset, we can then take a pair $(x_0^w, x_0^l)$ representing a winner and a loser, such that by using their associated scores they comply with $(x_0^w \succ x_0^l | c)$ where c is the conditional signal corresponding to the motion description. This process thus identify which motion within a pair better reflects the textual description.
With this dataset it is possible to achieve:
\begin{itemize}
    \item Increased Pairwise Comparisons: By generating $N=2K$ motions ($K$ per used model), we achieve up to $\binom{N}{2}$ potential winner-loser pairs per prompt. A high number of pairs is crucial for alignment algorithms.
    \item Preserved Multimodality: The high number of pairs associated with each prompt helps maintain multimodality. This prevents the alignment process from homogenizing the generated motions, reducing issues such as mode collapse.
\end{itemize}

\subsection{Motion Alignment Pipeline}
\label{sub4:mot_alig}
Our alignment pipeline is built upon \cite{wallace2023diffusion}, which is applicable to both DDPM and latent DDPM for motion synthesis.
The core idea of DPO is to optimize a conditional distribution \( p_\theta(x_0|c) \), instead of a reward model \( r(c, x_0) \) as done for RLHF approaches, while regularizing the KL-divergence from a reference distribution \( p_\text{ref} \):

\begin{equation}
\max_{p_\theta} \mathbb{E}_{c\sim \mathcal{D}_c,x_0\sim p_\theta(x_0|c)} \left[r(c,x_0)\right] - \beta \text{KL}\left[p_\theta(x_0|c)\|p_{\text{ref}}(x_0|c)\right]
\end{equation}

However, as stated by \cite{wallace2023diffusion}, the primary challenge in DPO for diffusion models is that the parameterized distribution $p_\theta(x_0|c)$ is not tractable,
as it needs to marginalize out all possible diffusion paths $(x_1, ..., x_T)$ which lead to $x_0$. With a few tricks and approximations this leads to an objective defined on the whole denoising path $x_{0:T}$:
\begin{equation}\label{eq:initial_dpo_loss}
    L_\text{DDPO}(\theta) = -\mathbb{E}_{x^w_{0},x^l_0}
    \log \sigma \left( \beta
    {\mathbb{E}}_{{\substack{x^w_{1:T} \sim p_\theta(x^w_{1:T}|x^w_0) \\ x^l_{1:T} \sim p_\theta(x^l_{1:T}|x^l_0)} }} 
    \left[ \log \frac{p_{\theta}(x^w_{0:T})}{p_\text{ref}(x^w_{0:T})} - \log \frac{p_{\theta}(x^l_{0:T})}{p_\text{ref}(x^l_{0:T})}\right]\right)\
\end{equation}

Intuitively, during training, the model samples points along the diffusion path and compares the newly generated data points with the preferred and non-preferred elements of a paired dataset.
The model is trained to improve its ability to denoise the \textit{winner} motion data points compared to the \textit{loser} ones, which are discouraged, according to the used ranking model, effectively aligning its generations with AI preferences.
Consequently, the training strategy guides the model towards generating motions that effectively match the desired characteristics defined by the used textual descriptions.
This alignment pipeline requires two models: a reference $p_{ref}$ and a target $p_{\theta}$ model that are identical, except for the key difference of reference model's parameters being completely frozen (i.e., not updated during training). 
We employ different tuning strategies depending on the type of model being used: for latent diffusion models, we only fine-tune denoisers' parameters. This approach is efficient because it operates directly in the latent space, eliminating the need to tune the entire diffusion network, which is also useful not to let the whole model shift too much from its previous behavior.
On the contrary, MDM denoises the raw input sequence, which is computationally inefficient, reason why we only fine-tune the last layers of the diffusion network.

\section{Experimental Evaluation}

This section discusses the used evaluation metrics in Sec.~\ref{sub5:evalmethods} and provides a quantitative (Sec.~\ref{sub5:quant}) and qualitative (Sec.~\ref{sub5:qual}) evaluation of our framework, and presents the results of ablation studies conducted to assess the contribution of different components to the overall performance.

\subsection{Evaluation Metrics}
\label{sub5:evalmethods}
We evaluate the generated motion sequences employing all metrics adopted in text-to-motion literature. The R-precision assesses the model's proficiency in producing relevant motion sequences based on textual queries. The Matching Score (MM-Dist) quantifies the alignment between the generated motion sequences and the textual descriptions used to generate them. Diversity measures the variability among the generated motion sequences, while Multi-Modality gauges the model's capacity to produce, starting from the same textual description, motion sequences across diverse modalities and styles.
The Fréchet Inception Distance (FID) \cite{heusel2017gans} measures the similarity between the distribution of generated motions and the distribution of ground truth. FID is a well-established metric \cite{brock2018large,dhariwal2021diffusion}, especially in image synthesis, which has been shown to be more consistent with the human judgment of image quality compared to the Inception Score (IS) \cite{salimans2016improved} and is used as a main performance indicator by capturing both diversity and fidelity to the ground truth dataset.
These metrics collectively provide a comprehensive evaluation of our model's performance in text-conditioned motion generation.

\subsection{Quantitative Evaluation}
\label{sub5:quant}
We conducted an extensive ablation study, described in Sec. \ref{sec6}, using MLD as base model and HumanML3D \cite{Guo22humanml3d} as main dataset.
Results on HumanML3D dataset are available in Table \ref{tab:humanml3d}, in which we reported results on the two rankers used to gather synthetic preferences as described in Sec. \ref{sub4:dataset} for both MDM and MLD.\newline
The best hyperparameter set, obtained from the ablation study on HumanML3D, is then used to further tune models on the KIT dataset, results depicted in Table \ref{tab:kit}.
Notice that we re-trained from scratch MLD on KIT due to the unavailability of the author's model.
Therefore, for a fairer comparison with the alignment approach we applied, in Table \ref{tab:kit} the results reported for MLD are those related to the model we trained from scratch. In addition, the only ranker we were able to use for KIT is backed by Guo's motion and text encoders due to an incompatible data representation used by TMR.\newline
It is worth mentioning that in all cases we obtain FID improvements while also retaining Multi-Modality and RPrecision Top3 performances. This is important especially because during alignment it may happen that tuned models can be affected by mode-collapse, thing that do not occur as we can infer by the improvement of FID and the solidity of Multi-Modality.

\begin{table}
\caption{Results for HumanML3D \cite{Guo22humanml3d}. Each group results corresponds to vanilla models, aligned models and current SOTA backed by a RAG framework. \textit{Italic} is for best results overall, while \textbf{bold} represents best results among pretrained models and MoDiPO aligned ones.
}
\label{tab:humanml3d}
\centering
\resizebox{0.8\textwidth}{!}{
{\renewcommand{\arraystretch}{1.2}
\begin{tabular}{c|c|c|c|c|c}
\toprule
\textbf{Model} & \textbf{Top3} $\uparrow$ & \textbf{MMDist} $\downarrow$ & \textbf{Diversity} $\rightarrow$ & \textbf{FID} $\downarrow$ & \textbf{MM} $\uparrow$ \\ \midrule
Real & $ .797^{\pm .002} $ & $ 2.974^{\pm .008} $ & $ 9.503^{\pm .065} $ & $ .002^{\pm .0} $ & -  \\ \midrule \midrule
MotionGPT \cite{jiang2024motiongpt} & $ .778^{\pm .002} $ & $ \mathit{3.096}^{\pm .008} $ & $ \mathit{9.528}^{\pm .071} $ & $ .232^{\pm .008} $ & $ 2.008^{\pm .084} $ \\
T2M-GPT \cite{zhang2023t2mgpt} & $ .775^{\pm .002} $ & $ 3.118^{\pm .011} $ & $ 9.761^{\pm .081} $ & $ \mathit{.116^{\pm .004}} $ & $ 1.856^{\pm .011} $ \\
MotionDiffuse \cite{zhang2022motiondiffuse} & $ \mathit{.782}^{\pm .001} $ & $ 3.113^{\pm .001} $ & $ 9.410^{\pm .049} $ & $ .630^{\pm .001} $ & $ 1.553^{\pm .042} $ \\ \midrule
MLD \cite{chen2023mld} & $ .755^{\pm .003} $ & $ 3.292^{\pm .01} $ & $ 9.793^{\pm .072} $ & $ .459^{\pm .011} $ & $ 2.647^{\pm .11} $ \\
Ours MLD-TMR & $ \mathbf{.758^{\pm .002}} $ & $ \mathbf{3.267^{\pm .01}} $ & $ 9.747^{\pm .077} $ & $ .303^{\pm .007} $ & $ 2.663^{\pm .111} $ \\
Ours MLD-GUO & $ .753^{\pm .003} $ & $ 3.294^{\pm .010} $ & $ \mathbf{9.702^{\pm .075}} $ & $ \mathbf{.281^{\pm .007}} $ & $ \mathbf{2.736^{\pm .111}} $ \\
\midrule
MDM \cite{tevet2023mdm} & $.703^{\pm .005}$ & $ 3.658^{\pm .025} $ & $ 9.546^{\pm .066} $ & $.501^{\pm .037}$ & $ \mathbf{2.874^{\pm .070}} $ \\
Ours MDM-TMR & $ \mathbf{.706^{\pm .004}} $ & $ \mathbf{3.634^{\pm .026}} $ & $ 9.531^{\pm .073} $ & $ \mathbf{.451^{\pm .031}} $ & $ 2.822^{\pm .069} $ \\
Ours MDM-GUO & $ .704^{\pm .005} $ & $ 3.641^{\pm .025} $ & $ \mathbf{9.495^{\pm .071}} $ & $ .486^{\pm .031} $ & $ 2.832^{\pm .066} $ \\ \midrule \midrule
ReMoDiffuse \cite{zhang2023remo} & $ .795^{\pm .004} $ & $ 2.974^{\pm .016} $ & $ 9.018^{\pm .075} $ & $ .103^{\pm .004} $ & $ 1.795^{\pm .043} $ \\ \bottomrule
\end{tabular}}
}
\end{table}
\begin{table}
\caption{
Results for KIT-ML \cite{plappert2016kit}. Each group results corresponds to vanilla models, aligned models and current SOTA backed by a RAG framework. \textit{Italic} is for best results overall, while \textbf{bold} represents best results among pretrained models and MoDiPO aligned ones.
}
\label{tab:kit}
\centering
\resizebox{0.8\textwidth}{!}{
{\renewcommand{\arraystretch}{1.2}
\begin{tabular}{c|c|c|c|c|c}
\toprule
\textbf{Model} & \textbf{Top3} $\uparrow$ & \textbf{MMDist} $\downarrow$ & \textbf{Diversity} $\rightarrow$ & \textbf{FID} $\downarrow$ & \textbf{MM} $\uparrow$ \\ \midrule
Real & $ .779^{\pm .006} $ & $ 2.788^{\pm .012} $ & $11.08^{\pm .097} $ & $ .031^{\pm .004} $ & - \\ \midrule \midrule
MotionGPT \cite{jiang2024motiongpt} & $ .680^{\pm .005} $ & $ 3.527^{\pm .021} $ & $10.35^{\pm .084} $ & $ \mathit{.510}^{\pm .016} $ & $ \mathit{2.328}^{\pm .117} $ \\
T2M-GPT \cite{zhang2023t2mgpt} & $ \mathit{.745}^{\pm .006} $ & $ 3.007^{\pm .023} $ & $10.92^{\pm .108} $ & $ .514^{\pm .029} $ & $ 1.570^{\pm .039} $ \\
MotionDiffuse \cite{zhang2022motiondiffuse} & $ .739^{\pm .004} $ & $ \mathit{2.958}^{\pm .005} $ & $\mathit{11.10}^{\pm .143} $ & $ 1.954^{\pm .062} $ & $ 0.730^{\pm .013} $ \\ \midrule
MLD \cite{chen2023mld} & $ .718^{\pm .005} $ & $ \mathbf{3.156^{\pm .023}} $ & $10.921^{\pm .108} $ & $ .733^{\pm .032} $ & $ 1.595^{\pm .087} $ \\
Ours MLD-GUO& $ \mathbf{.720^{\pm .006}} $ & $ 3.166^{\pm .03} $ & $ \mathbf{11.022^{\pm .074}} $ & $ \mathbf{.623^{\pm .037}} $ & $ \mathbf{1.660^{\pm .068}} $ \\
\midrule
MDM \cite{tevet2023mdm} & $ \mathbf{.726^{\pm .005}} $ & $ 3.098^{\pm .023} $ & $ \mathbf{10.690^{\pm .088}} $ & $ .496^{\pm .028} $ & $ \mathbf{1.829^{\pm .037}} $ \\
Ours MDM-GUO & $ 0.725^{\pm .005} $ & $ \mathbf{3.097^{\pm .023}} $ & $10.681^{\pm .086} $ & $ \mathbf{.450^{\pm .026}} $ & $ 1.821^{\pm .038} $ \\ \midrule \midrule
ReMoDiffuse \cite{zhang2023remo} & $ .765^{\pm .055} $ & $ 2.814^{\pm .012} $ & $10.80^{\pm .105} $ & $ .155^{\pm .006} $ & $ 1.239^{\pm .028} $ \\ \bottomrule
\end{tabular} }
}
\end{table}

MoDiPO's main purpose is to discourage the generation of implausible motions by improving consistency of diffusion-based text-to-motion models. While these models can generate impressive motions, the stochasticity inherent in the diffusion process can sometimes lead to the generation of poor-quality motions, either affected by foot-skating or jittering, or incoherent with the input prompt.\newline
Aiming to show that models tuned with our MoDiPO methodology are more consistent even in the case of multiple generations for the same prompt, we synthesize eight different motions for each prompt of HumanML3D test set. We then ranked these motions using the Guo Ranker, which is also used for dataset synthetic annotation and ablation study as described in Sec. \ref{sub6:rank_methods}. Finally, we compute all the used evaluation metrics exclusively for the worst and best motion sequences from each set of eight generations.\newline
As shown in Table \ref{tab:modiopo_rationale_a} for HumanML3D and Table \ref{tab:modiopo_rationale_b} for KIT-ML, the vanilla MLD model exhibits a significant difference in FID scores between the worst and best motions. By contrast, using MoDiPO, models significantly reduce the difference between the FID scores for loser and winner motions. In particular, we observe a gap reduction of 15.05\% (from 0.711 to 0.604) for HumanML3D and of 96.51\% (from 0.172 to 0.006) for KIT-ML.
It is worth noting that the FID score for losers decreases from 0.997 to 0.789 (an improvement of 20.86\%) for HumanML3D and from 1.037 to 0.769 (an improvement of 25.84\%) for KIT-ML.
Besides the reduced gap between loser and winner, the lower overall FID for the aligned model suggests that MoDiPO is a practical solution for improving the consistency of motion quality in diffusion models.

\begin{table}
\caption{Performances of Winner (W) and Loser (L) among the 8 generations per-prompt of vanilla MLD (MLD) and MLD aligned with MoDiPO (Ours) for HumanML3D and KIT-ML. The typical evaluation method calculation is used in the case of MLD and Ours (aligned model).}
\begin{minipage}[t]{0.51\textwidth}
\begin{subtable}{\textwidth}
\caption{HumanML3D}
\label{tab:modiopo_rationale_a}
\centering
\resizebox{\textwidth}{!}{
\begin{tabular}{c|c|c|c|c}
\toprule
\textbf{Model} & \textbf{Top3} $\uparrow$ & \textbf{MMDist} $\downarrow$ & \textbf{Div.} $\rightarrow$ & \textbf{FID} $\downarrow$ \\ \midrule
Real & $.797^{\pm .002}$ & $2.974^{\pm .008}$ & $~9.503^{\pm .065}~$ & $.002^{\pm .0}$ \\ \midrule
MLD & $.755^{\pm .003}$ & $3.292^{\pm .010}$ & $9.793^{\pm .072}$ & $.459^{\pm .011}$ \\
MLD L & $.511^{\pm .003}$ & $4.822^{\pm .014}$ & $9.249^{\pm .079}$ & $.997^{\pm .026}$ \\
MLD W & $.914^{\pm .001}$ & $2.341^{\pm .004}$ & $9.942^{\pm .075}$ & $.286^{\pm .007}$ \\
\midrule
Ours & $.753^{\pm .003}$ & $3.294^{\pm .010}$ & $9.702^{\pm .075}$ & $.281^{\pm .007}$ \\
Ours L & $.487^{\pm .002}$ & $4.945^{\pm .014}$ & $9.056^{\pm .069}$ & $.789^{\pm .021}$ \\
Ours W & $.916^{\pm .001}$ & $2.303^{\pm .007}$ & $9.873^{\pm .079}$ & $.185^{\pm .005}$ \\ \bottomrule
\end{tabular}
}
\end{subtable}
\end{minipage}
\hfill
\begin{minipage}[t]{0.53\textwidth}
\begin{subtable}{\textwidth}
\caption{KIT-ML}
\label{tab:modiopo_rationale_b}
\centering
\resizebox{\textwidth}{!}{
\begin{tabular}{c|c|c|c|c} \toprule
\textbf{Model} & \textbf{Top3} $\uparrow$ & \textbf{MMDist} $\downarrow$ & \textbf{Div.} $\rightarrow$ & \textbf{FID} $\downarrow$ \\ \midrule
Real & $.779^{\pm .006}$ & $2.788^{\pm .012}$ & $~11.080^{\pm .097}~$ & $.031^{\pm .004}$ \\
\midrule
MLD & $.718^{\pm .005}$ & $3.156^{\pm .023}$ & $10.921^{\pm .108}$ & $.733^{\pm .032}$ \\
MLD L & $.619^{\pm .007}$ & $4.001^{\pm .034}$ & $10.829^{\pm .100}$ & $1.037^{\pm .053}$ \\
MLD W & $.799^{\pm .006}$ & $2.672^{\pm .024}$ & $11.302^{\pm .115}$ & $.865^{\pm .039}$ \\
\midrule
Ours & $.720^{\pm .006}$ & $3.166^{\pm .030}$ & $11.022^{\pm .074}$ & $.623^{\pm .037}$ \\
Ours L & $.615^{\pm .004}$ & $4.048^{\pm .026}$ & $10.805^{\pm .106}$ & $.769^{\pm .047}$ \\
Ours W & $.802^{\pm .005}$ & $2.646^{\pm .022}$ & $11.297^{\pm .114}$ & $.763^{\pm .034}$ \\ \bottomrule
\end{tabular}
}
\end{subtable}
\end{minipage}
\end{table}

\subsection{Qualitative Evaluation}
\label{sub5:qual}
\begin{figure}
    \centering
    \begin{subfigure}[b]{.49\textwidth}
        \centering
        \begin{subfigure}[b]{.49\textwidth}
            \includegraphics[width=\textwidth]{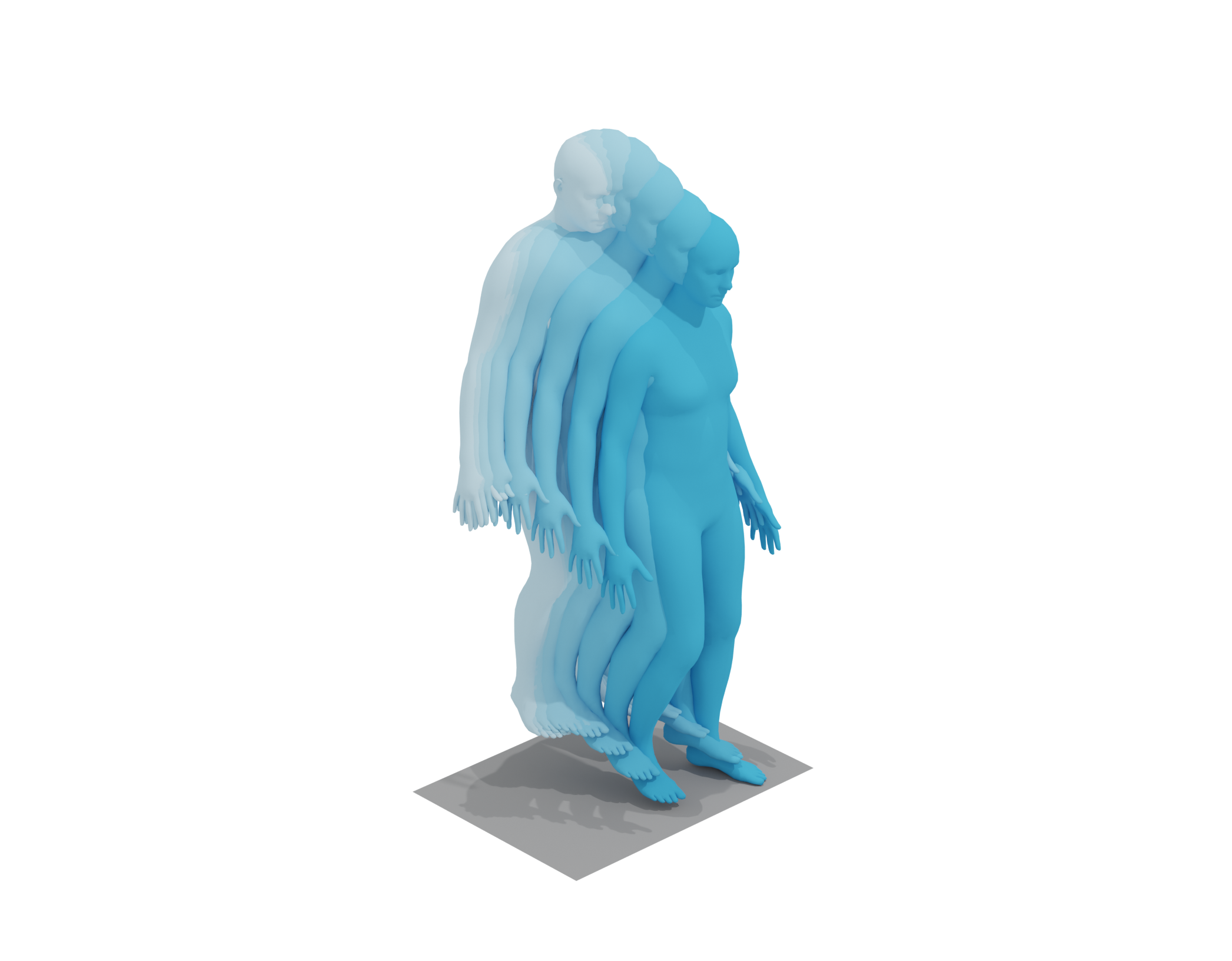}
            \label{fig:image1}
        \end{subfigure}
        \begin{subfigure}[b]{.49\textwidth}
            \includegraphics[width=\textwidth]{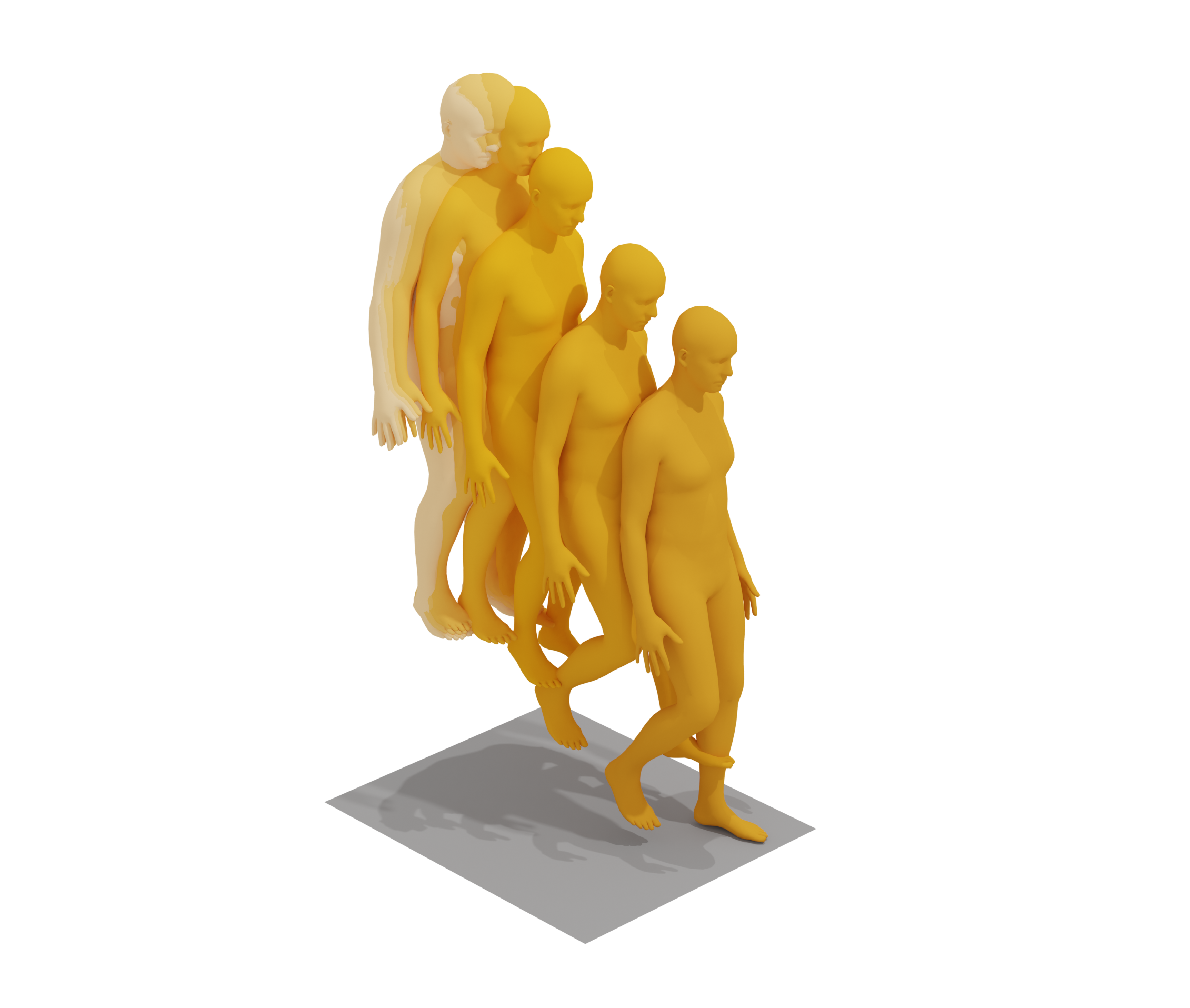}
            \label{fig:image2}
        \end{subfigure}
        \caption{A person walks down the stairs}
        \label{fig:quala}
    \end{subfigure}
    \hfill
    \begin{subfigure}[b]{.49\textwidth}
        \centering
        \begin{subfigure}[b]{.49\textwidth}
            \includegraphics[width=\textwidth]{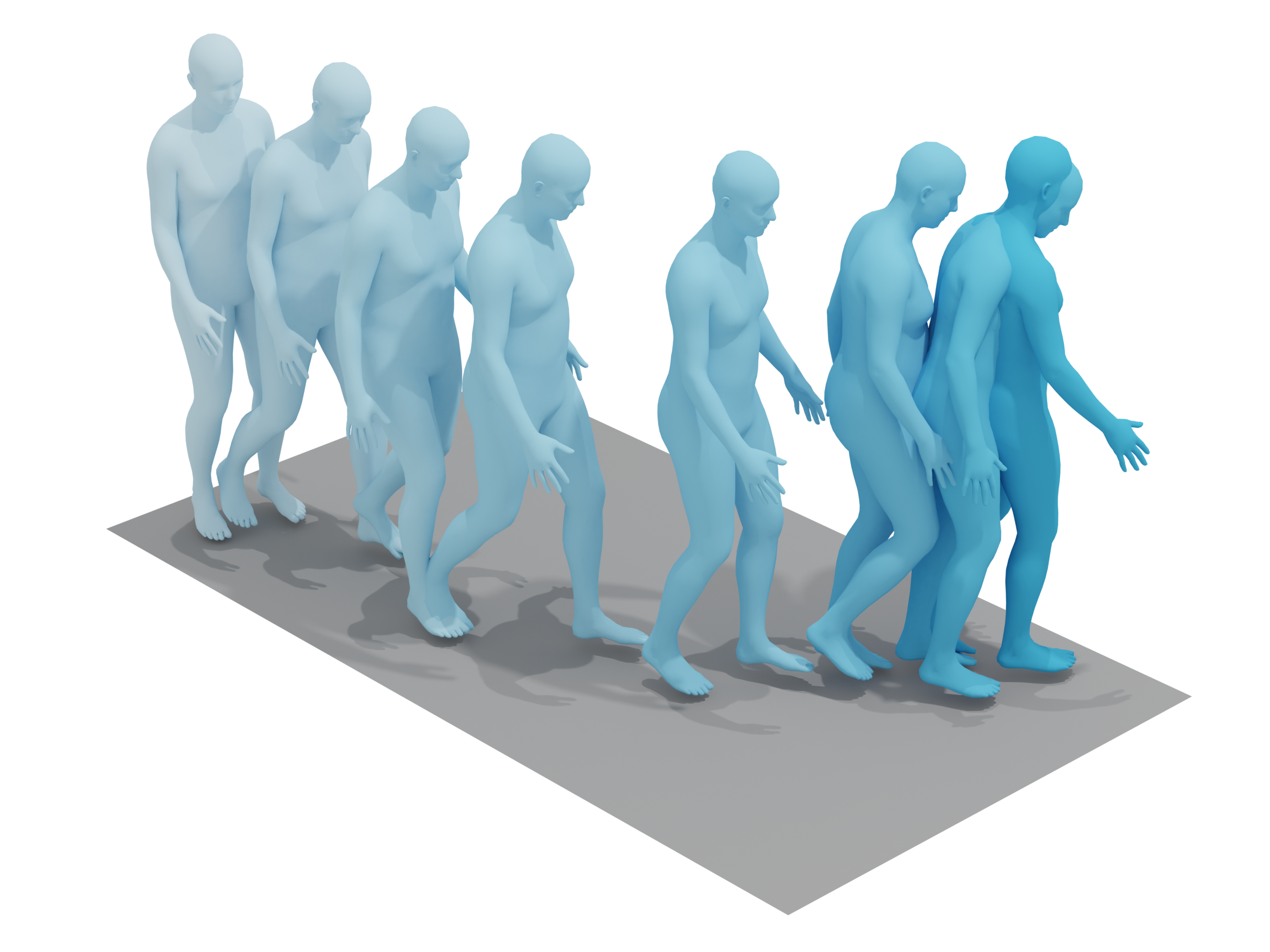}
            \label{fig:image5}
        \end{subfigure}
        \begin{subfigure}[b]{.49\textwidth}
            \includegraphics[width=\textwidth]{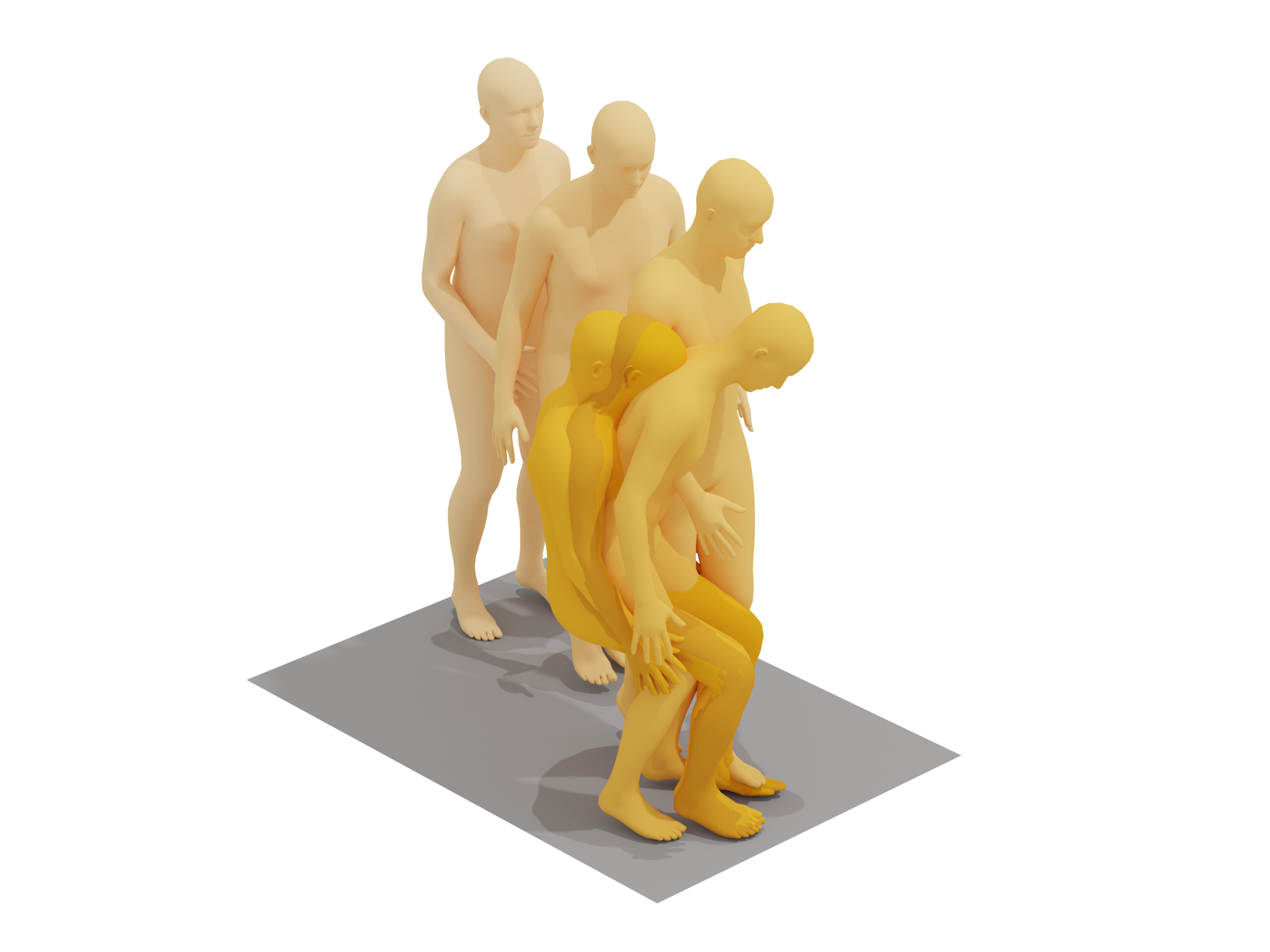}
            \label{fig:image6}
        \end{subfigure}
        \caption{A person walks forwards, sits}
        \label{fig:qualb}
    \end{subfigure}
    
    \begin{subfigure}[b]{.49\textwidth}
        \centering
        \begin{subfigure}[b]{.49\textwidth}
            \includegraphics[width=\textwidth]{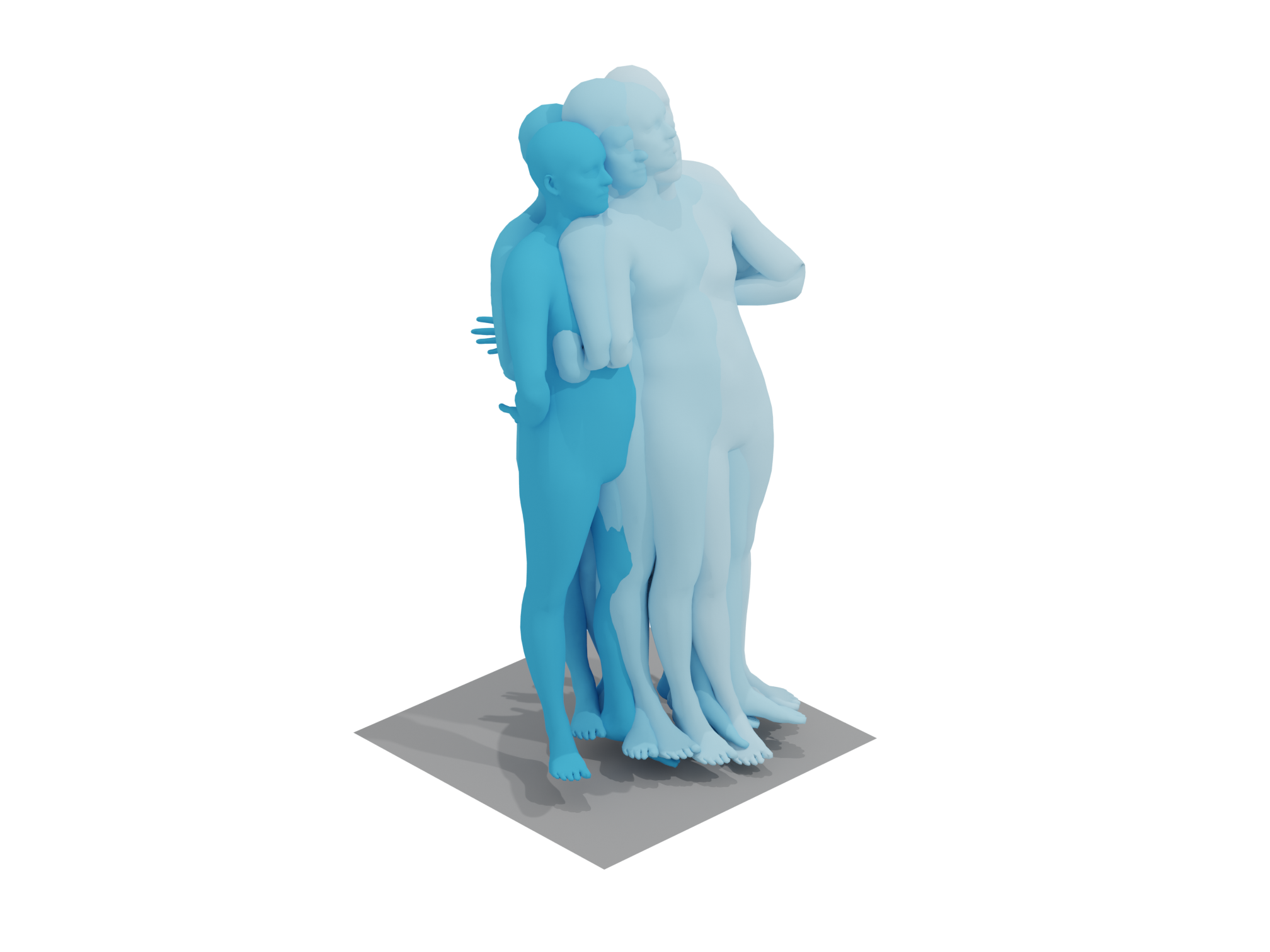}
            \label{fig:image7}
        \end{subfigure}
        \begin{subfigure}[b]{.49\textwidth}
            \includegraphics[width=\textwidth]{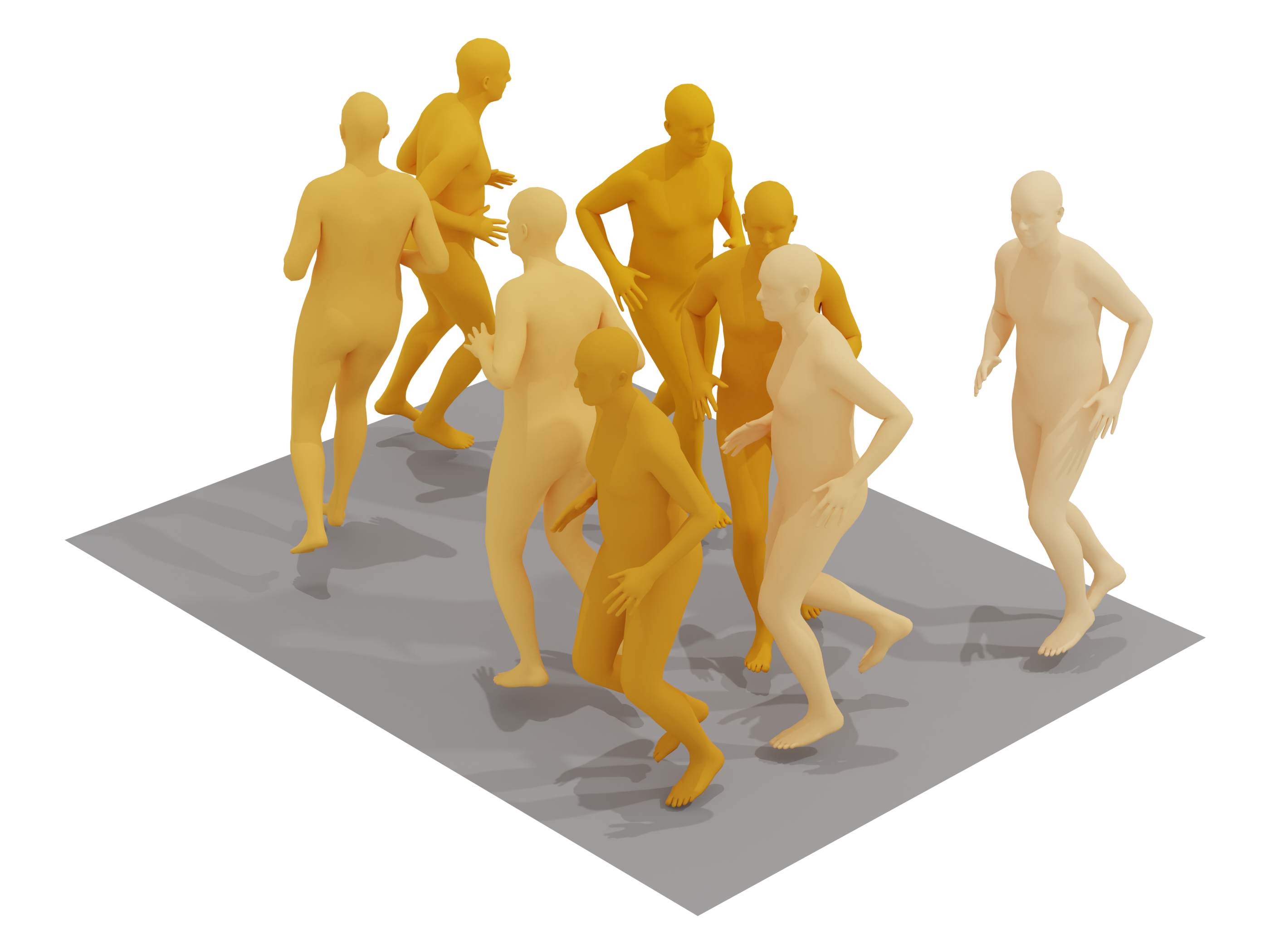}
            \label{fig:image8}
        \end{subfigure}
        \caption{A person is jogging around}
        \label{fig:qualc}
    \end{subfigure}
    \begin{subfigure}[b]{.49\textwidth}
        \centering
        \begin{subfigure}[b]{.49\textwidth}
            \includegraphics[width=\textwidth]{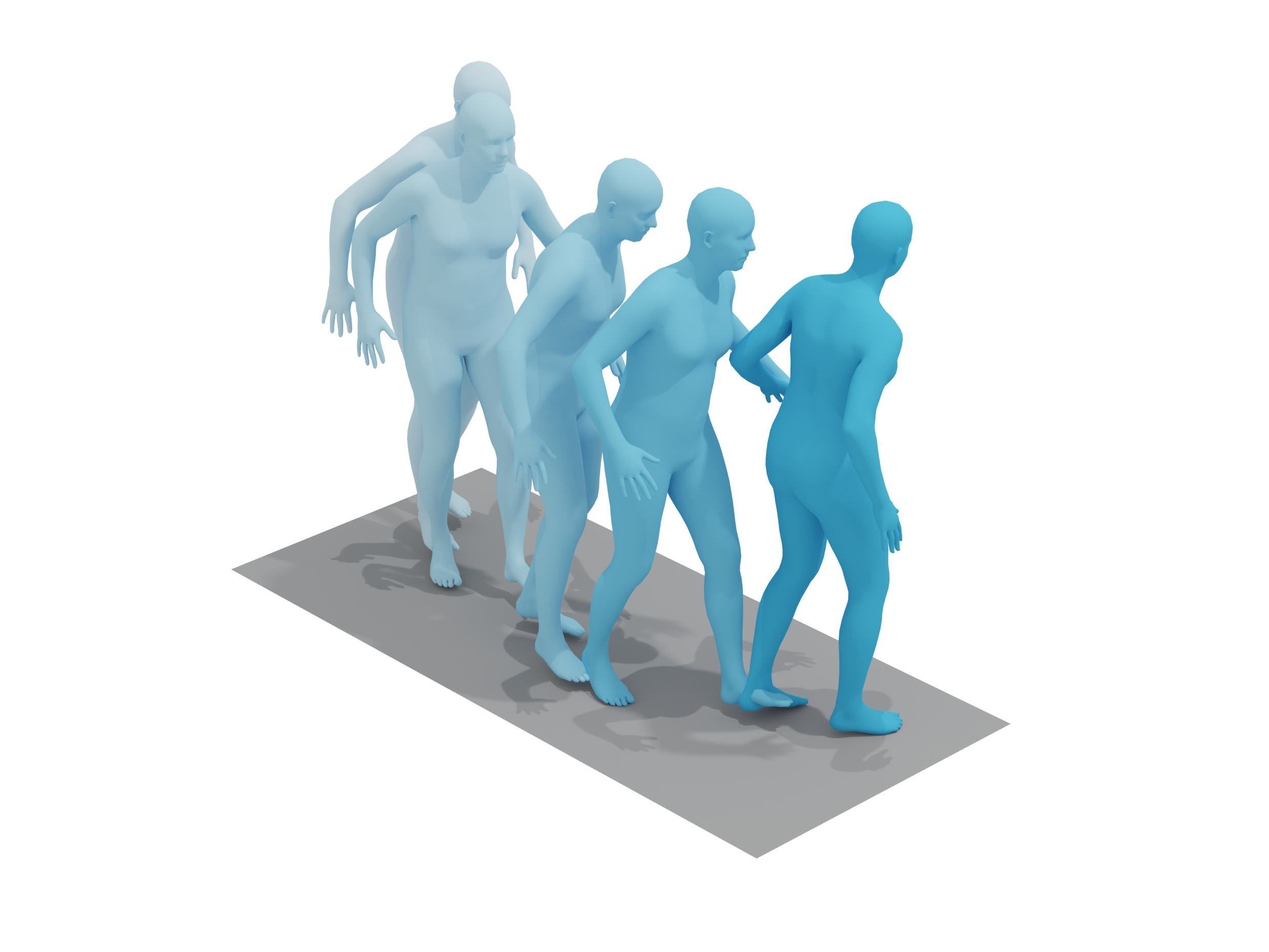}
            \label{fig:image9}
        \end{subfigure}
        \begin{subfigure}[b]{.49\textwidth}
            \includegraphics[width=\textwidth]{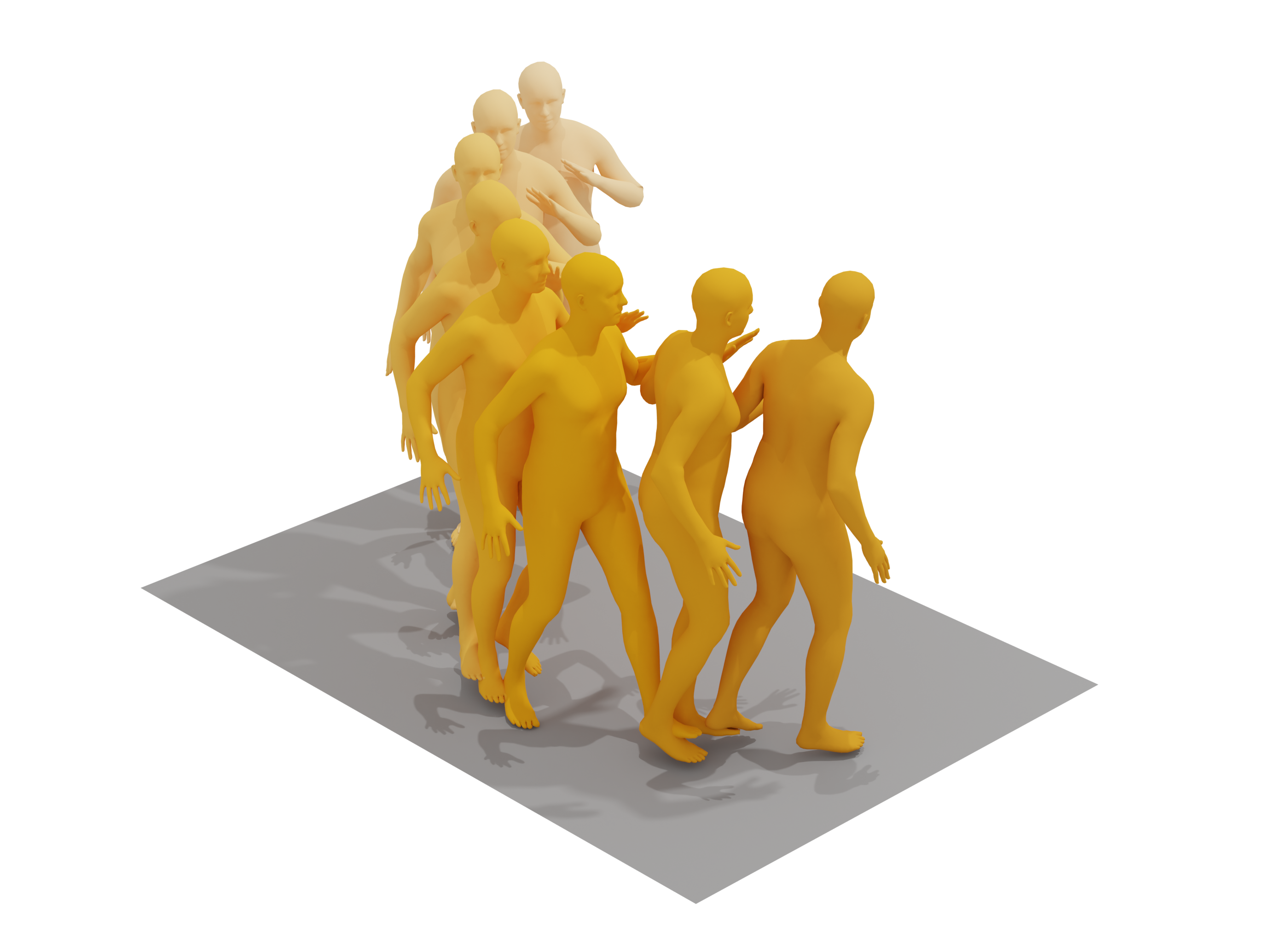}
            \label{fig:image10}
        \end{subfigure}
        \caption{A person is walking around in circles with their left arm on or in something}
        \label{fig:quald}
    \end{subfigure}
    
    \begin{subfigure}[b]{.49\textwidth}
        \centering
        \begin{subfigure}[b]{.49\textwidth}
            \includegraphics[width=\textwidth]{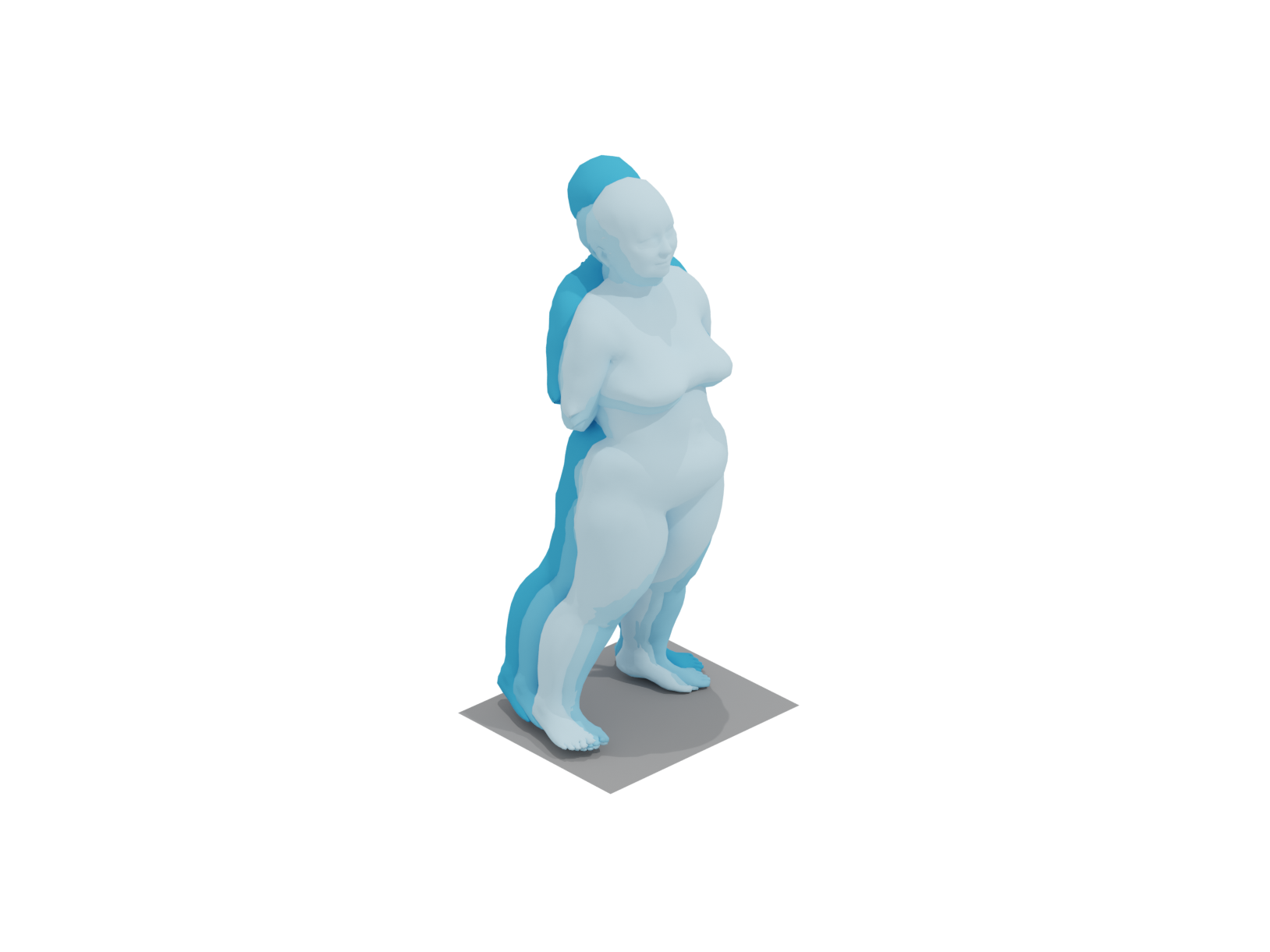}
            \label{fig:image13}
        \end{subfigure}
        \begin{subfigure}[b]{.49\textwidth}
            \includegraphics[width=\textwidth]{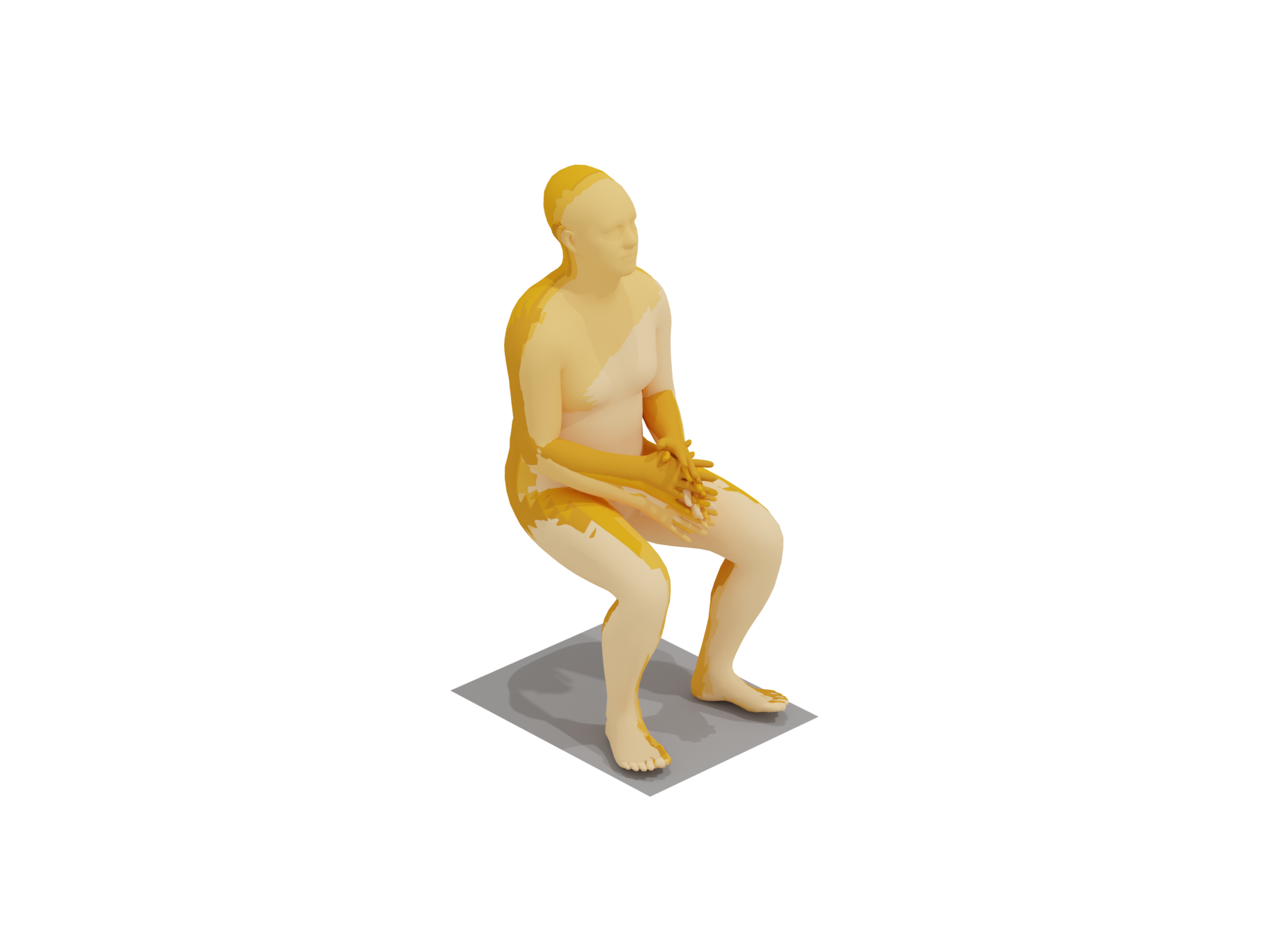}
            \label{fig:image14}
        \end{subfigure}
        \caption{A person is sitting with both hands together}
        \label{fig:quale}
    \end{subfigure}
    \hfill
    \begin{subfigure}[b]{.49\textwidth}
        \centering
        \begin{subfigure}[b]{.49\textwidth}
            \includegraphics[width=\textwidth]{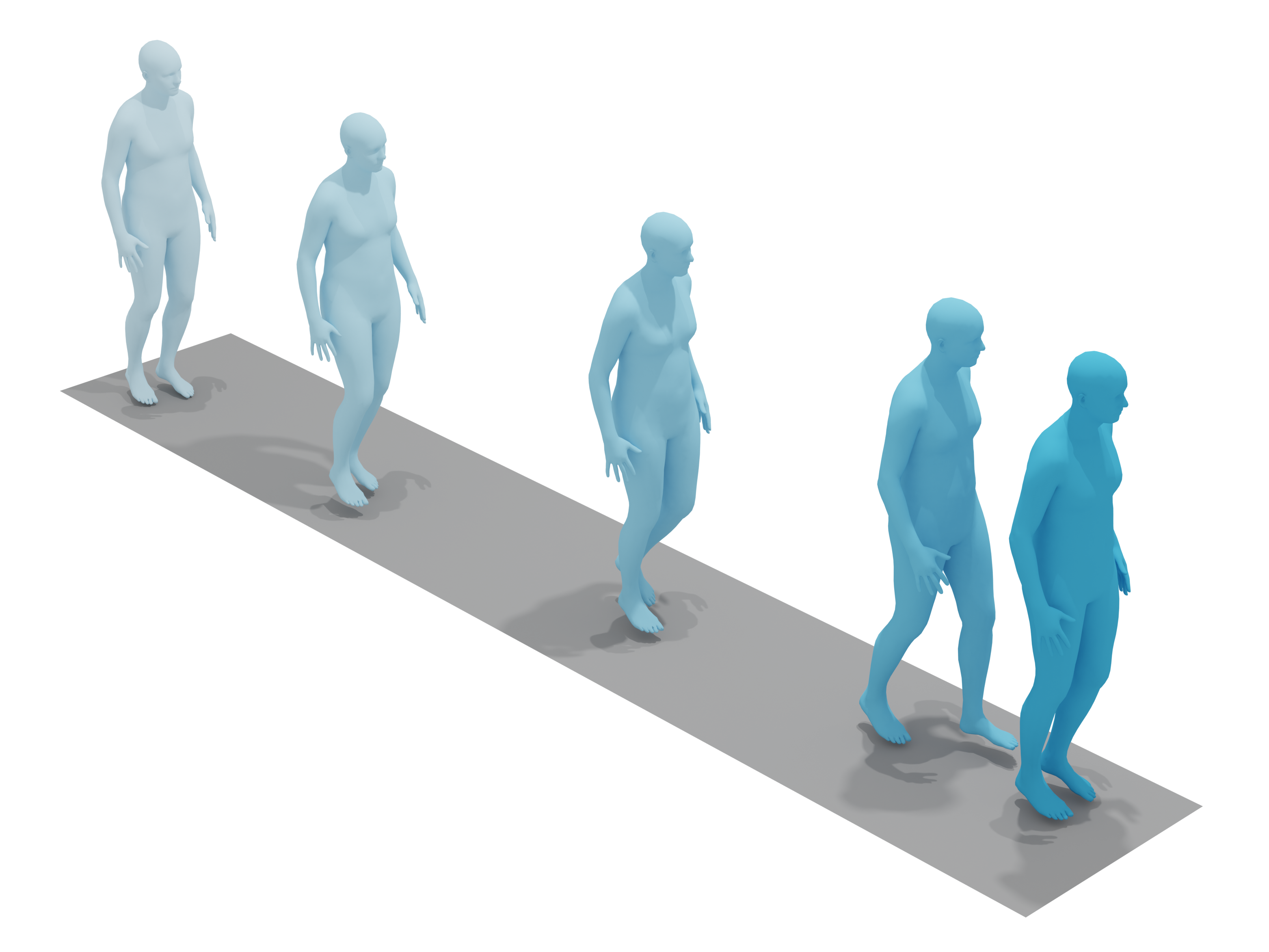}
            \label{fig:image15}
        \end{subfigure}
        \begin{subfigure}[b]{.49\textwidth}
            \includegraphics[width=\textwidth]{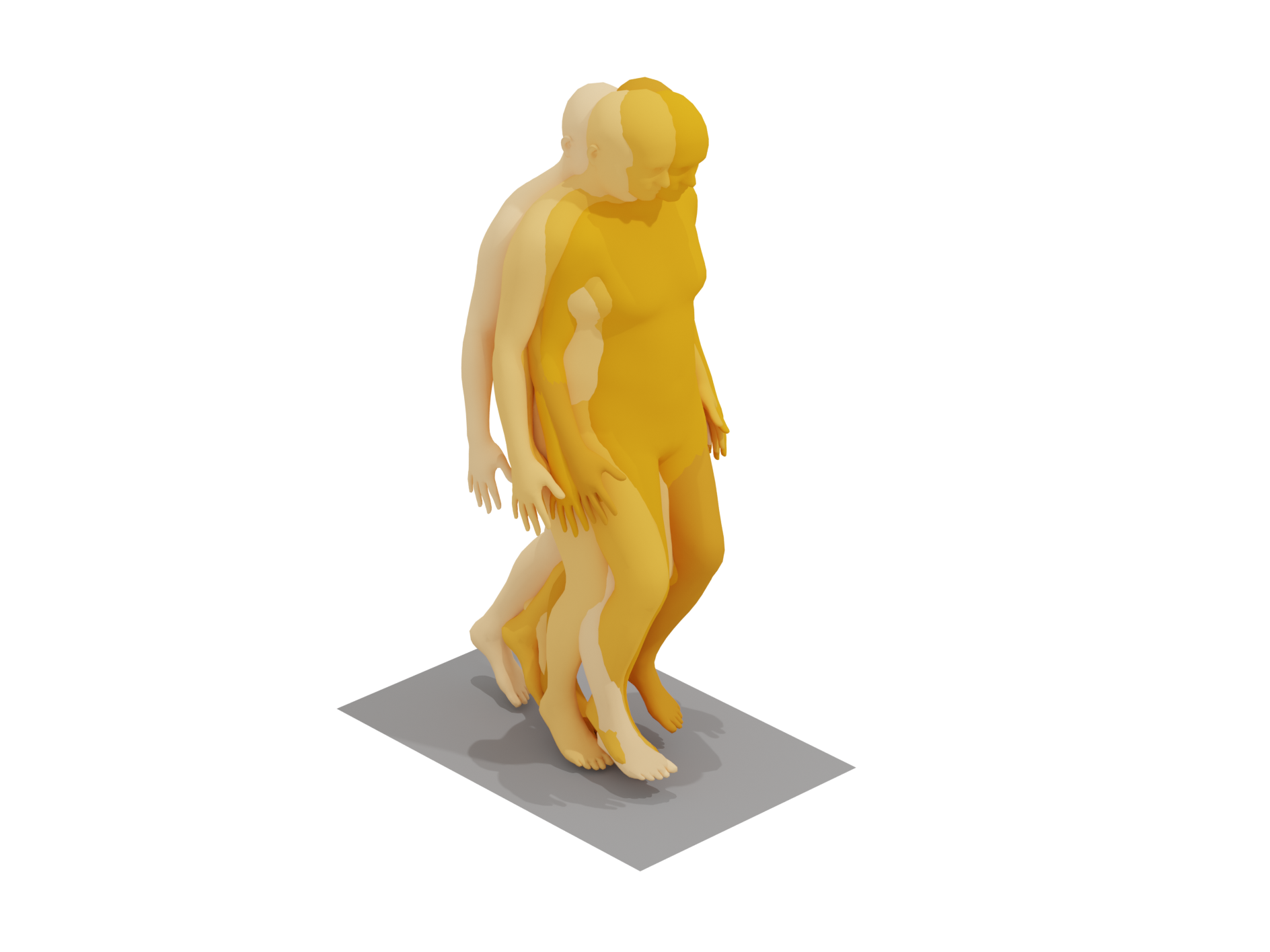}
            \label{fig:image16}
        \end{subfigure}
        \caption{A person is jogging forward in a steady manner}
        \label{fig:qualf}
    \end{subfigure}

    \caption{Qualitative results on HumanML3D. Vanilla MLD is represented by blue motions, while MLD aligned with MoDiPO is represented by yellow motions.}
    \label{fig:qualitatives}
\end{figure}

To perform qualitative evaluation we generated motions using the best performing model from our ablation study (Table \ref{tab:ablation}).
In Fig. \ref{fig:qualitatives} it is possible to see qualitatives of both pre-trained MLD and the post-DPO version of MLD.
Even though it is difficult to judge movements just from pictures, it is easy to see the improvement from these examples. In particular from the prompt \textit{"a person walks down the stairs"} of Fig.~\ref{fig:quala}, the tuned model generates a motion that is more aligned with a "walk down the stairs" action.\newline
At the same time, in the case of Fig.~\ref{fig:qualb} with the prompt \textit{"A person walks forwards, sits"}, the post-DPO model better capture the "sits" movement.\newline
In Fig.~\ref{fig:qualc} for vanilla-MLD model the \textit{"jogging"} action is performed on the spot, while the aligned model is more coherent to the used prompt and to a more realistic execution of movements described by it.\newline
However, in Fig.~\ref{fig:qualf} it is possible to see how tricky the input prompt can be: in this case the pre-trained model generated a motion of a person walking forward, while the post-DPO model seems to give more importance to the \textit{"steady"} part of the description, which thus outputs a motion of a person steady walking, a movement made on the spot.

\section{Ablation Studies}
\label{sec6}
We conducted a series of ablation studies to analyze the impact of specific choices within the proposed method.
In Sec. \ref{sub6:sampling}, we investigated the effects of using stochastic or edge sampling of winner-loser pairs for the ranking process.
Sec. \ref{sub6:rank_methods} covers the evaluation of performances for different ranking methods used within the framework.
In Sec. \ref{sub6:own_or_wother}, there is a description of the advantages and disadvantages of using motions generated by models other than the model we aim to align.
Finally, Sec. \ref{sub6:online} illustrates an online alignment experiment.

\subsection{Pair Selection}
\label{sub6:sampling}
With the dataset described in Sec. \ref{sub4:dataset}, we choose two strategies of winner-loser selection: edge and stochastic selection.
In the edge selection, we just take hard winners and hard losers from the ranked list of motion sequences.\newline
For the stochastic selection, instead, we divided the list into two parts of which the first half contains winners, while the second half includes losers.
According to our experiments, as can be seen in Table~\ref{tab:ablation}, this second strategy performs better if compared to the edge selection.\newline
We suppose that the model, subjected to different generations for the same textual prompt, retains its R-Precision and Multi-Modality performances while also reducing the FID thanks to alignment with DPO.\newline
We also tried to substitute the winner, with a probability of $0.25$ or $0.5$, with the ground truth motion as additional supervision to further enhance the possibility of generating more realistic motions. However, as depicted in Table~\ref{tab:ablation}, even if this setting helps decreasing FID and increasing Multi-Modality, it deteriorates RPrecision performances.

\subsection{Ranking Method}
\label{sub6:rank_methods}
As described in Sec. \ref{sub4:dataset}, we propose to synthetically rank generated motions starting from two models: TMR and the motion and text encoders proposed in \cite{Guo22humanml3d}, which we will dub as Guo. While TMR has been proposed to tackle retrieval tasks, Guo was introduced for evaluation purposes.
For both the used datasets, KIT and HumanML3D, we annotated each set composed of text motions with Guo. Instead, TMR was used only for HumanML3D annotation. This is mainly due to a mismatch in the KIT data representation, which was updated to the SMPL-like representation in the case of TMR. This made it impossible for us to use it in KIT synthetic ranking, knowing that the motions generated by MLD and MDM have a parameter mismatch with this updated representation.
Results of experiments done with both the used rankers are available in Table \ref{tab:ablation}.

\subsection{Alignment from model's own generations}
\label{sub6:own_or_wother}
As described in Sec. \ref{sub4:dataset}, we used synthetically annotated datasets generated by MLD and MDM models. We hypothesized that incorporating motions generated by MDM during MLD tuning could offer potential benefits. Indeed, a greater amount of winner-loser pairs help maintain multimodality and R-Precision performance, as previously observed.
At the same time, utilizing motions from both models could expose the target model to different approaches for generating good motions based on the textual prompt.
However, as shown in Table \ref{tab:ablation}, this approach did not yield the expected improvements. The performance of the target model decreases if compared to using only its own generated motions for tuning.

\subsection{Online Alignment}
\label{sub6:online}
We explored an alternative approach using online alignment with the MLD model. This approach is feasible due to MLD's latent-space operation, which significantly reduces the time required to generate output motions.\newline
In this scenario, the frozen reference model generates motions which are then ranked and from which we selected winner-loser pairs using the stochastic selection. This entire process occurs online, offering potential benefits in terms of efficiency and adaptability.\newline
However, initial results show a trade-off between different metrics. While the target model's FID decreases, suggesting improved quality in distribution similarity, the R-Precision Top3 metric also decreases.
This suggests that for each winner-loser pair more iterations are necessary to let the target model effectively repulse loser motions and attract winner ones.

\begin{table}
\caption{Ablation study with MLD model on HumanML3D dataset \cite{Guo22humanml3d}. Stoch stands for stochastic selection of pairs, default is edge selection. $GT_p$ is the probability that there is supervision with the ground truth sequence used at the spot of winner in a pair. Online refers to the generation and ranking, on the fly, of sequences generated by the reference model. PaM+ stands for Pick-a-Move dataset used in its entirety, meaning that generations of both MLD and MDM are used for the winner-loser pair selection. Reported evaluation metrics are RPrecision Top3, MMDistance, Diversity, FID and Multi-Modality.}
\label{tab:ablation}
\centering
\resizebox{0.9\textwidth}{!}{
\begin{tabular}{c|c|c|c|c|c|c|c|c|c}
Scorer & Stoch.    & $GT_p$ & Online         & PaM$+$      & Top3 $\uparrow$         & MMDist $\downarrow$   & Diversity $\rightarrow$             & FID $\downarrow$     & MM $\uparrow$ \\
\hline
\multicolumn{5}{c|}{without alignment}        & $ .755^{\pm .003} $    & $ 3.292^{\pm .010} $    & $ 9.793^{\pm .072} $   & $ .459^{\pm .011} $   & $ 2.647^{\pm .110} $  \\
\hline
Guo  & $\checkmark$  &      &               &                & $ .753^{\pm .003} $    & $ 3.294^{\pm .010} $   & $ 9.702^{\pm .075} $   & $ .281^{\pm .007} $   & $ 2.736^{\pm .111} $  \\
TMR   & $\checkmark$  &      &               &                & $ .758^{\pm .002} $    & $ 3.267^{\pm .010} $   & $ 9.747^{\pm .077} $   & $ .303^{\pm .007} $   & $ 2.663^{\pm .111} $  \\
Guo  &               &      &               &                & $ .752^{\pm .003} $    & $ 3.300^{\pm .010} $   & $ 9.714^{\pm .077} $   & $ .336^{\pm .009} $   & $ 2.736^{\pm .110} $  \\
TMR   &               &      &               &                & $ .759^{\pm .002} $    & $ 3.262^{\pm .010} $   & $ 9.714^{\pm .079} $   & $ .325^{\pm .007} $   & $ 2.630^{\pm .107} $  \\
Guo  & $\checkmark$  & .25  &               &                & $ .745^{\pm .002} $    & $ 3.324^{\pm .010} $   & $ 9.674^{\pm .077} $   & $ .261^{\pm .007} $   & $ 2.834^{\pm .119} $  \\
Guo  & $\checkmark$  & .5   &               &                & $ .745^{\pm .003} $    & $ 3.327^{\pm .010} $   & $ 9.614^{\pm .076} $   & $ .273^{\pm .007} $   & $ 2.826^{\pm .117} $  \\
Guo  & $\checkmark$  &      & $\checkmark$  &                & $ .677^{\pm .003} $    & $ 3.701^{\pm .013} $   & $ 9.241^{\pm .079} $   & $ .276^{\pm .008} $   & $ 3.477^{\pm .144} $  \\
Guo  & $\checkmark$  & .25  & $\checkmark$  &                & $ .714^{\pm .003} $    & $ 3.496^{\pm .010} $   & $ 9.489^{\pm .074} $   & $ .293^{\pm .009} $   & $ 3.179^{\pm .128} $  \\
Guo  & $\checkmark$  &      &               &  $\checkmark$  & $ .761^{\pm .003} $    & $ 3.252^{\pm .010} $   & $ 9.851^{\pm .071} $   & $ .335^{\pm .008} $   & $ 2.610^{\pm .110} $  \\
TMR   & $\checkmark$  &      &               &  $\checkmark$  & $ .759^{\pm .002} $    & $ 3.266^{\pm .010} $   & $ 9.798^{\pm .073} $   & $ .345^{\pm .007} $   & $ 2.592^{\pm .104} $
\end{tabular}
}
\end{table}

\section{Limitations}
While our study introduces the first alignment framework for text-driven generative models for human motion generation, there are several limitations that require consideration. In our approach, we fine-tune the entire denoiser for Motion Latent Diffusion (MLD) and the last layers of Motion Diffusion Model (MDM), which may not be optimal as it could underperform if compared to other fine-tuning strategies such as LoRA \cite{lora}, or fine-tuning the whole model. In the future, we aim to employ parameter efficient fine-tuning (PEFT) strategies such as LoRA and its variants for more computation-intensive models such as MDM.

\section{Conclusions}
We have presented the first framework designed for aligning text-to-motion generative models. Our approach significantly enhanced the aesthetics of motion generation, as demonstrated through both qualitative and quantitative analyses, showcasing improvements of up to $39\%$ on the Fréchet Inception Distance (FID). Utilizing AI-feedback, we constructed a preferential dataset, augmenting its diversity by introducing stochasticity in the selection process of winner-loser pairs. Leveraging the training dataset enabled us to explore innovative strategies, such as incorporating ground-truth motion, further enhancing the model's adherence to the underlying data distribution.

\bibliographystyle{splncs04}
\bibliography{main}
\end{document}